\documentclass{bmvc2k}
\usepackage{amsfonts}

\usepackage{color}
\definecolor{turquoise}{cmyk}{0.65,0,0.1,0.1}
\definecolor{purple}{rgb}{0.65,0,0.65}
\definecolor{dark_green}{rgb}{0, 0.4, 0}
\definecolor{dark_blue}{rgb}{0, 0, 0.4}
\definecolor{orange}{rgb}{0.6, 0.3, 0.0}
\definecolor{red}{rgb}{0.8, 0.2, 0.2}
\definecolor{brown}{rgb}{0.5, 0.16, 0.16}

\newcommand{\rui}[1]{{\color{black}#1}}

\newcommand{\rev}[1]{{\color{black}#1}}
\newcommand{\revsupp}[1]{{\color{black}#1}}

\graphicspath{{../pdf/}{./figs/}}

\title{Spatial-Temporal Residual Aggregation for High Resolution Video Inpainting}

\addauthor{Vishnu Sanjay Ramiya Srinivasan$^\ast$}{vishnusanjay.rs@gmail.com}{1}
\addauthor{Rui Ma$^\ast$}{ruim@jlu.edu.cn}{2,1}
\addauthor{Qiang Tang$^\dagger$}{qiang.tang@huawei.com}{1} 
\addauthor{Zili Yi}{yizili14@gmail.com}{1}
\addauthor{Zhan Xu}{zhan.xu@huawei.com}{1}

\addinstitution{
 Huawei Technologies\\
 Canada
}
\addinstitution{
 Jilin University\\
 China
}

\runninghead{Srinivasan and Ma et al.}{Spatial-Temporal Residual Aggregation}

\begin{document}

\maketitle

\vspace{-20pt}

\begin{abstract}

Recent learning-based inpainting algorithms have achieved compelling results for completing missing regions after removing undesired objects in videos.
To maintain the temporal consistency among the frames, 3D spatial and temporal operations are often heavily used in the deep networks.
However, these methods usually suffer from memory constraints and can only handle low resolution videos.
We propose STRA-Net, a novel spatial-temporal residual aggregation framework for high resolution video inpainting.
The key idea is to first learn and apply a spatial and temporal inpainting network on the downsampled low resolution videos. 
Then, we refine the low resolution results by aggregating the learned spatial and temporal image residuals (details) to the upsampled inpainted frames. 
Both the quantitative and qualitative evaluations show that we can produce more temporal-coherent and visually appealing results than the state-of-the-art methods on inpainting high resolution videos.

\end{abstract}
\vspace{-14pt}
\section{Introduction}
\label{sec:intro}
\vspace{-6pt}

Video inpainting is a long-studied task in computer vision and has wide applications such as video restoration \cite{chang2008application}, removing undesired objects \cite{abraham2012survey, ilan2015survey} etc.
Although numerous efforts, including the traditional optimization based \cite{bertalmio2001navier,wexler2007space, newson2014video, huang2016temporally} and the more recent deep learning based techniques \cite{chang2019free,chang2019learnable,kim2019deep,lee2019copy,kim2019recurrent,zeng2020learning,xu2019deep,gao2020flow,zou2021progressive}, have been paid to tackle this challenge, most of the existing methods can only handle low resolution videos due to various reasons, e.g., memory or computation time constraints and the lack of high resolution training data.

\begin{figure}
	\centering
	\includegraphics[width=0.75\linewidth]{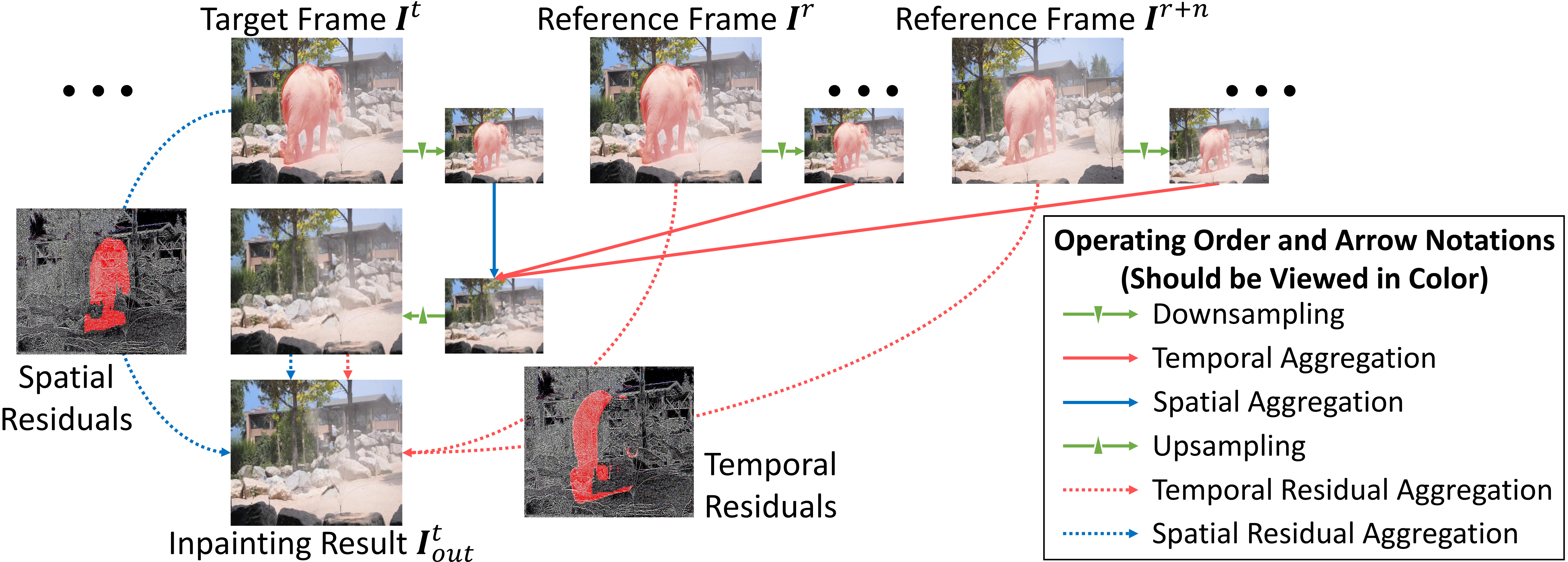}	
	\vspace{-8pt}
	\caption{Given a video sequence (top row) with masks of undesired objects (red region), we first perform spatial and temporal aggregation on the downsampled low resolution frames. Then, we employ spatial and temporal residual aggregation for transferring residuals (details) to the upsampled impainted image and generate the high resolution inpainting result. 
	}
	\label{fig:teaser}
	\vspace{-15pt}
\end{figure}

In this paper, we aim for \textit{high resolution} video inpainting without additional memory constraints and expect the inpainting network to be trained only on the low resolution videos.
To this end, we propose STRA-Net, a novel spatial-temporal residual aggregation framework for high resolution video inpainting (Figure \ref{fig:teaser}).
The key idea is to first learn and apply a spatial-temporal inpainting network (STA-Net) on the downsampled low resolution videos. 
Then, we refine the low resolution results by aggregating the learned spatial and temporal \textit{image residuals }(i.e., high frequency details) to the upsampled inpainted frames. 

The STRA-Net is inspired by recent advance in high resolution image inpainting which employs the contextual (or spatial) residual aggregation \cite{yi2020contextual} to recover the details of the inpainted region using the intra-frame information.
Besides the spatial residual aggregation, we also introduce a novel \textit{temporal residual aggregation} module to transfer image details from other neighboring frames. 
The STRA-Net is modularized, while the only trainable module STA-Net just needs to be trained and inferred on the low resolution videos.
Hence, our method is memory efficient and not constrained by the resolution of the input video.

We conduct quantitative and qualitative comparisons with the state-of-the-art methods for high resolution video inpainting.
From the experiments, our method can produce more temporal-coherent and visually appealing results.
We also conduct ablation studies to verify the effectiveness of the proposed new modules.
In summary, our contributions are as follows:
\begin{itemize}
	\itemsep0pt
	\vspace{-4pt} 
	\item We propose STRA-Net, a novel spatial-temporal residual aggregation framework that enables high resolution video inpainting without the memory constraints or requirements of high resolution training data.
	\item We combine a novel temporal residual aggregation module with the spatial residual aggregation to transfer high frequency image details from inter and intra video frames.
	\item We propose multiple new modules in STA-Net, such as a more robust joint image alignment module and the multi-scale spatial and temporal aggregation, to improve the performance of the learning-based video inpainting pipeline.
\end{itemize}
\vspace{-12pt}
\section{Related Works}
\vspace{-2pt}
Video inpainting is a long-studied yet still active research area \cite{abraham2012survey, ilan2015survey}.
In this section, we mainly focus on the deep-learning based video inpainting techniques which have achieved compelling results recently.
In addition, as a related topic, we summarize the latest methods for high resolution image inpaiting.

\vspace{-4pt}
\subsection{Deep Video Inpainting}
\vspace{-2pt}

A common solution for video inpainting is to complete the missing content using the spatial and temporal information gathered from the current or other video frames.
In the deep learning regime, one typical practice for learning the spatial and temporal information in videos is to employ 3D convolutions to fuse the spatial-temporal features across multiple frames.
Constrained by the GPU memory, inpainting methods employing 3D convolutions \cite{chang2019free,chang2019learnable,wang2019video,huproposal} and recurrent propogation \cite{kim2019recurrent,ding2019frame} can only work on low resolution videos.

Another way to aggregate the spatial and temporal information is to compute the image alignment between frames and use the image correspondences to transfer information.
Based on how the image alignment is estimated, most of recent techniques can be classified into either learning-based or flow-based methods.
The learning-based methods \cite{lee2019copy,oh2019onion,li2020short,zeng2020learning,liu2021decoupled} directly learn the alignment transformations from the pairwise image features.
Afterwards, the missing regions can be predicted based on all aligned images, allowing information transferred even from temporal distant frames.
However, such methods also often suffer from the memory constraints and require high resolution data for training.
On the other hand, the flow-based methods \cite{chang2019vornet,zhang2019internal, xu2019deep,gao2020flow,zou2021progressive} mainly perform pixel-wise data propagation between consecutive frames with the guidance of optical flow \cite{ilg2017flownet, hur2019iterative}. 
The flow-based methods have shown promising results for inpainting high resolution videos, but they may produce artifacts when the flow estimation is not accurate, e.g., for videos containing fast or delicate motion. 
In addition, optical flow mostly detects image correspondences in the short temporal range and struggles with long temporal range pixel alignment \cite{li2020short}.
Therefore, in this paper, we explore the possibility of combining strengths from both the learning-based and flow-based methods for a more robust joint image alignment.
Besides performing the temporal aggregation using the proposed image alignment, we also employ the spatial aggregation which is proposed in the image inpainting \cite{yu2018generative, yi2020contextual} to refine each inpainted frame.



\vspace{-8pt}
\subsection{High Resolution Image Inpainting}
\vspace{-2pt}
For video inpainting, there may occur some scenarios where no reference pixels can be found from other frames. 
An image inpainting step is usually incorporated into a video inpainting pipeline to fill the remaining missing regions with some visually plausible content.
Recently, deep image inpainting methods \cite{pathak2016context,iizuka2017globally,yang2017, yu2018generative,zeng2019PENnet,yi2020contextual} have achieved impressive results by using techniques such as GAN \cite{goodfellow2014} and attention based content transfer. 
For example, Yu et al. \cite{yu2018generative} proposed a contextual attention based inpainting method which can transfer features from distant spatial locations outside the hole region.
In our spatial aggregation module, we use multi-scale contextual attention transfer to further refine the leftover region in which the pixels are synthesized by the network instead of copied from the reference frames. 

While most of recent methods work effectively for low resolution images, they don't generalize well to high resolution images due to memory constraints or the requirement of high resolution training data.
To overcome these limitations, Yi et al. \cite{yi2020contextual} introduced the contextual residual aggregation (CRA) based algorithm HiFill to inpaint ultra high resolution images with less memory usage.
Image residuals, which encode the high frequency details in the image, have been widely studied in low-level vision tasks, such as Laplacian pyramids construction \cite{burt1987laplacian,toet1989image}, edge detection \cite{sharifi2002classified} and image quality assessment \cite{deshmukh2010image}.
In \cite{yi2020contextual}, the residual images computed by a simple yet effective downsampling and upsampling process are aggregated using the contextual attention transfer and then added to the low resolution inpainting result. 
In the end, a sharp high resolution inpainted image can be generated.
In STRA-Net, we extend the idea of CRA to the temporal domain and transfer residuals from multiple temporal reference images to improve the low resolution video inpainting result.

\vspace{-6pt}
\section{Method}

\begin{figure*}
	\centering
	\includegraphics[width=0.84\linewidth]{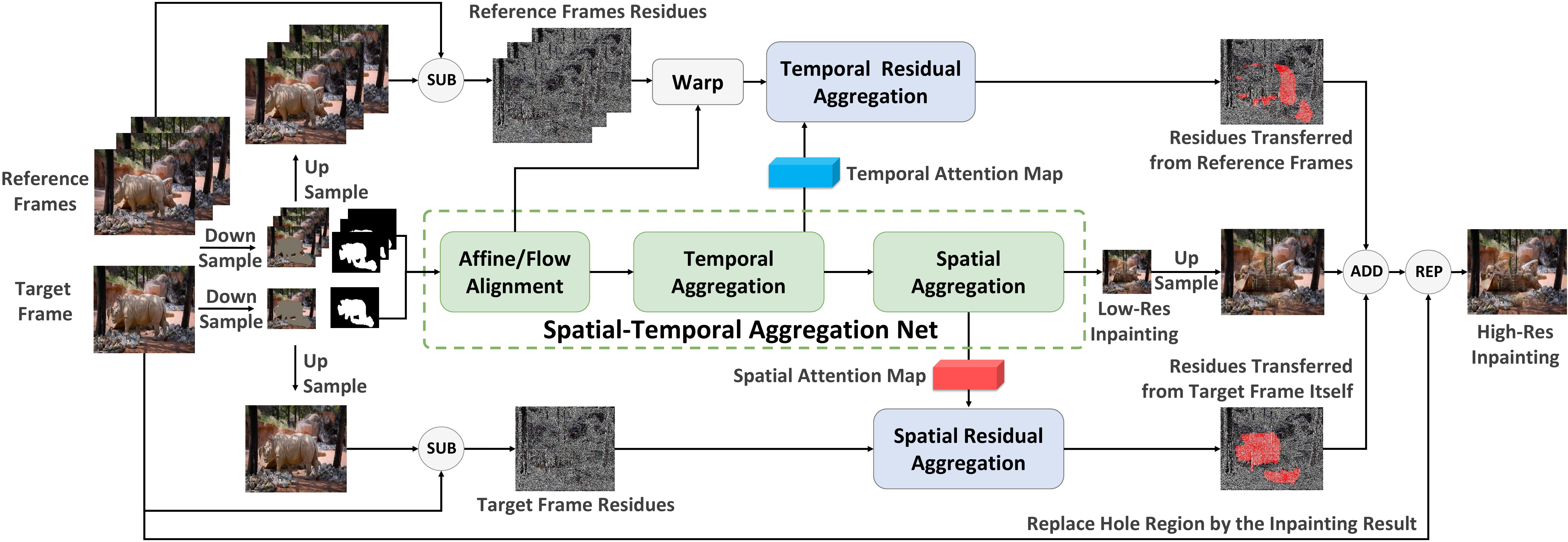}
		\vspace{-6pt}
	\caption{Overview of the STRA-Net. The STA-Net (highlighted by green) learns to inpaint the downsampled low resolution videos.
	The high frequency image details are added to the upsampled low resolution results by the spatial and temporal residual aggregation.
}
	\label{fig:overview}
		\vspace{-15pt}
\end{figure*}


\vspace{-4pt}
\subsection{Overview of STRA-Net}
\vspace{-4pt}
The overview of our spatial-temporal residual aggregation framework (STRA-Net) is shown in Figure \ref{fig:overview}. 
The system takes a high resolution video with the masks of undesired regions (holes) as input and outputs an inpainted high resolution video with the holes filled. 
The video is processed frame-by-frame in the temporal order. 
We refer the frame to be filled as the \textit{target frame} and other frames as the \textit{reference frames}.
First, all the raw frames and masks are padded and downsampled to a lower resolution. 
Then, we learn a spatial-temporal aggregation network (STA-Net) to inpaint the low resolution target frame.
To generate the residual images which contain the high frequency details, we upsample the previously downsampled frames and masks, and subtract them from the raw inputs. 
In the end, the spatial and temporal residual aggregation modules use the spatial and temporal attention maps computed from the STA-Net to generate a sharp inpainted frame at the original resolution.
The inpainted frame will subsequently be used as a reference frame, providing more information for filling the following frames. 

\vspace{-6pt}
\subsection{STA-Net}
\vspace{-2pt}
The STA-Net aims to inpaint the target video frame using the spatial and temporal information from the reference frames.
Due to the robustness and extensibility of the learning-based method, we build our network structure based on a general learning-based pipeline, Copy-Paste Network (CPNet) \cite{lee2019copy}.
In this section, we first provide a summary of CPNet and then introduce the STA-Net modules which improve the CPNet in all aspects. 

\vspace{2pt}
\noindent \textbf{CPNet summary.}
CPNet is a two-stage pipeline which contains three modules.
In the first stage, an Alignment network is trained to compute affine transformation to align the reference frames to the target frame.
Pixel-wise $L1$ distance is minimized between the target frame and the aligned reference frame on regions where the pixels are \textit{visible} (non-hole) in both the frames.
%
In the second stage, an encoder-decoder structure is used to extract and aggregate the features of the target and reference frames (Copy network), and reconstruct the target frame from the aggregated feature (Paste network).
A key component in the Copy network is a \textit{context matching module} which computes the features that can be copied from the aligned reference frames to the target frame.
First, a cosine similarity $C_{r,t}$ is calculated between the target feature map $\mathbf{F}^t$ and each aligned reference feature map $\mathbf{F}^{r \to t}$.
Then, a \textit{masked softmax} operation is applied to the computed feature similarities to generate a \textit{saliency map} (attention scores or weights) which is used to aggregate the features from each $\mathbf{F}^{r \to t}$ to $\mathbf{F}^t$.
In the Paste network, the target frame feature is concatenated with the aggregated reference features and fed to the decoder, which consists of several dilated convolution blocks and upsampling operations.  
The target frame is reconstructed in the decoder by either filling the holes if any aggregated reference is available or generating the pixels using dilated convolutions if no reference information is found in the previous step.
Here, we only briefly summarize the pipeline of CPNet and we ignore the operations on the masks for the simplicity of presentation. More details about CPNet can be found in \cite{lee2019copy}.  

\begin{figure*}
	\centering
	\includegraphics[width=0.85\textwidth]{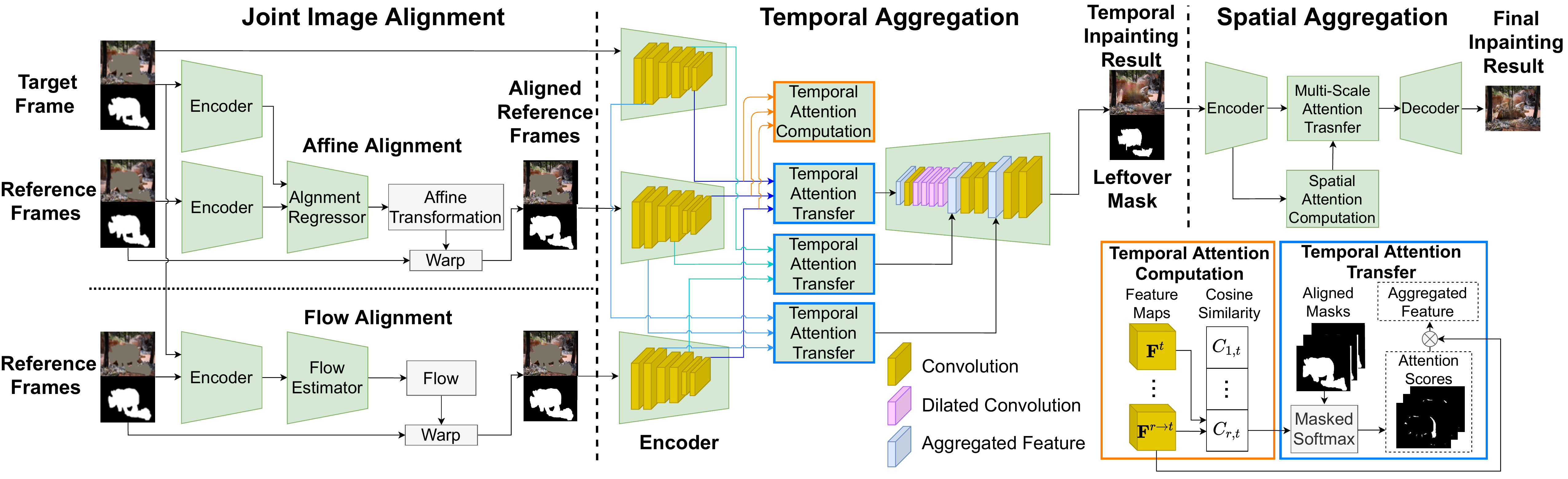}
	\vspace{-6pt}
	\caption{Overview of the STA-Net. 
		All trainable modules are in green.
	The image encoders have the same architecture composed of Convolution and MaxPool layers, but their weights are only shared within the same module.
	Each Temporal Attention Transfer module (blue) uses the same attention scores computed from the Temporal Attention Computation module (orange) to transfer features at different scales, while 
	 we omit to draw the connections between them to avoid the clutteredness.}
	\label{fig:sta-net}
			\vspace{-14pt}
\end{figure*}

\noindent \textbf{STA-Net pipeline.} We extend the CPNet to three stages and propose novel improvements on each stage (Figure \ref{fig:sta-net}).
In the first stage, we propose a joint alignment network which combines affine transformation and optical flow for better image alignment between the target and the reference frames.
Secondly, we apply a multi-scale temporal aggregation network to copy information from references frames and generate an initial inpaiting result.
Finally, we augment the CPNet with a new refinement stage which uses a spatial aggregation network to refine the result using contextual information from the target image itself.

\vspace{2pt}
\noindent \textbf{Joint alignment network.}
Computing a robust image alignment within a proper temporal window is crucial to copy valuable and relevant contents from the other frames.
The affine transformation based alignment \cite{lee2019copy}, which we refer to as \textit{affine alignment}, provides a large temporal window to copy information from distant frames and is best suited for static background based on our experiments.
In contrast, the optical flow based alignment \cite{xu2019deep,gao2020flow} (\textit{flow alignment} in short) has a small temporal range, but can handle local motion due to the dense image correspondences computed.
In our approach, we jointly employ the affine and flow alignment to generate two types of aligned reference frames (Figure \ref{fig:sta-net} left): one is computed by applying the learned affine transformation to each reference frame; the other is obtained by using the optical flow to propagate the pixels from the neighboring reference frames to the target frame.
The affine transformation branch is implemented similarly as in CPNet, while for the optical flow branch, we directly use a pre-trained PWC-IRR network \cite{hur2019iterative} to compute the optical flow.
The target frame and all two types of aligned references frames are passed to the temporal aggregation module.

\vspace{2pt}
\noindent \textbf{Multi-scale temporal aggregation network.}
After alignment, the target frame and aligned reference images are converted into multi-scale feature maps using a shared image encoder.
Motivated by the multi-scale attention transfer and score sharing for efficient spatial feature aggregation introduced in \cite{yi2020contextual}, we propose \textit{multi-scale temporal attention transfer} to aggregate the temporal features in a multi-scale manner.
This module can also be treated as a multi-scale extension of the contextual matching module in CPNet.
To increase the efficiency, we share the temporal attention scores calculated at the lowest resolution of the multi-scale feature maps and perform attention transfer on multiple higher resolution feature maps (Figure \ref{fig:sta-net} middle).
The attention scores are computed by applying a masked softmax function on the cosine similarity $C_{r,t}$ between the target feature $\mathbf{F}^t$ and the aligned reference frame feature $\mathbf{F}^{r \to t}$ (Figure \ref{fig:sta-net} bottom right).
This multi-scale feature aggregation scheme can ensure more refined feature selection when transferring the features from reference frames.
The temporally aggregated feature is then passed to a similar image decoder as in CPNet which outputs an inpainted image.
Note that for certain regions in the result, there may be no feature transferred from other frames.
Such regions correspond to a \textit{leftover mask} where the temporal attention scores are zero for all reference frames.
The pixels in the leftover mask are generated by the dilated convolutions in the decoder and thus may look blurry.
Next, we will refine the regions covered by the leftover mask using the spatial aggregation.

\vspace{2pt}
\noindent \textbf{Multi-scale spatial aggregation network.}
Given an initial inpainting result and a leftover mask, our goal is to refine the masked regions using the contextual information within the target image.
This task is actually similar to the detail refinement step in the existing image inpainting methods. 
Here, we employ the refinement network from HiFill \cite{yi2020contextual} as our multi-scale spatial aggregation module due to its efficiency and the ability to transfer the multi-scale features.
The spatial aggregation network has a similar encoder-decoder structure  to the temporal aggregation network, but operates in the spatial domain to copy multi-scale features from the contextual image patches in the target frame rather than the reference frames (Figure \ref{fig:sta-net} top right).
The spatial attention scores are computed based on pairs of contextual patches which contain one patch from the masked region (holes) and one patch from the non-hole regions.
Similar to the multi-scale temporal attention transfer, the spatial attention transfer also aggregates the target image features at multiple resolutions using the shared spatial attention scores. 
At last, the spatially aggregated feature is passed to the decoder which generates a refined inpainting result for the region covered by the leftover mask. 


\vspace{-6pt}
\subsection{Spatial-Temporal Residual Aggregation (STRA)}
\label{sec:st-res}
\vspace{-4pt}
The STA-Net can work well for inpainting low resolution videos, but may not handle high resolution videos due to the memory constraints.
One naive way to use STA-Net for high resolution video inpainting is to first downsample the input video and then upsample the inpainted result of STA-Net.
However, this process will generate blurry results due to the upsampling. 
Inspired by the contextual residual aggregation based algorithm HiFill for efficient high resolution image inpainting \cite{yi2020contextual}, 
we learn and apply STA-Net at the low resolution and then perform spatial and temporal residual aggregation to improve the inpainting result (Figure \ref{fig:overview}).
We extract the temporal and spatial residuals from the reference frames and the target frame by a regular subtraction process described earlier.
Then, the \rui{affine and flow alignment learned from the} STA-Net are used to align the temporal residual images to the target frame.
Next, we perform temporal residual aggregation by using the temporal attention scores computed in the temporal aggregation network  to transfer high frequency details from aligned temporal residual images to the upsampled STA-Net output.
Similarly, we perform spatial residual aggregation to transfer details from surrounding patches to further refine the inpainted region as in HiFill. 
In the end, we succeed to generate a sharp inpainted output at its original resolution.

\vspace{-6pt}
\section{Training}
\vspace{-4pt}
The only trainable module in our framework is the STA-Net which is trained end-to-end by randomly sampling five frames from a synthesized training dataset (at 512x512 resolution) and optimized with the losses proposed below.
In the inference stage, the network can take more than five frames (e.g., 20 in our implementation) to compute the image alginment and spatial-temporal aggregation across the frames.


\noindent \textbf{Training dataset.}
To obtain videos with inpainting ground truth, we follow a similar procedure as in \cite{lee2019copy}. First, we composited various object/human masks to the 1.8M images from Places \cite{zhou2017places}. Then, we generated the image sequences (\rui{5 frames each}) by random 2D image transformations.
Besides the originally used masks from Pascal VOC \cite{everingham2015pascal}, we also added masks from MS COCO \cite{lin2014microsoft} and Open Image \cite{kuznetsova2018open} to increase the number and diversity of the missing region masks.
In total, we have collected 335.7K masks.
\rui{To further increase the dataset, we also downloaded natural scene videos from the Internet, processed them into 33.7K video clips (10 frames each) and composited the object masks similarly as before.}
\rev{
All frames are resized and padded so that the final training data is at 512x512.
}

\vspace{2pt}
\noindent \textbf{Losses.} The total training loss for STA-Net is defined as follows:
\setlength{\abovedisplayskip}{1pt}
\setlength{\belowdisplayskip}{2pt}
\setlength{\jot}{0pt}
\begin{multline} \label{eq:total_loss}
L_{total} = 5 \cdot L_{align} + 10 \cdot L_{hole(visible)} + 20 \cdot L_{hole(leftover)}  +  6 \cdot L_{non-hole} \\ 
\vspace{-10pt}
+ 0.01 \cdot L_{perceptual} + 24 \cdot L_{style} + 1.2 \cdot L_{rec} + 0.001 \cdot L_{adv}.
\end{multline}
Here, $L_{total}$ is composed of two parts: one is similar to the losses defined in the CPNet for image alignment and temporal aggregation; the other employs the losses in HiFill to refine the leftover region by spatial aggregation. 
Specifically, 
$L_{align}$ measures the $L1$ distance between the aligned reference frames and the target frame, and is used to supervise the alignment network. 
$L_{hole(visible)}$, $L_{hole(leftover)}$ and $L_{non-hole}$ are $L1$ distances measured on the reconstructed images for the pixels lying in the temporal aggregated, the leftover or the non-hole regions, respectively.
Similar to CPNet, we apply perceptual and style losses, $L_{perceptual}$ and $L_{style}$, to further improve the visual quality of the results.
%
%
%
%
At last, following HiFill, we use the $L1$ distance based reconstruction loss $L_{rec}$ and the WGAN-GP loss \cite{gulrajani2017improved} as the adversarial loss $L_{adv}$ to train the spatial aggregation network.
\rev{The mathematical definition} and
more training details can be found in the supplementary material.

\begin{table*}
	\centering
	\footnotesize
	\addtolength{\tabcolsep}{-4pt}
	\begin{tabular}{l|ccc| ccc| ccc| ccc }
		&  \multicolumn{3}{c|}{432 $\times$ 240} &  \multicolumn{3}{c|}{864 $\times$ 480} &  \multicolumn{3}{c|}{1296 $\times$ 720} &  \multicolumn{3}{c}{1728 $\times$ 960} \\
		
		\multicolumn{1}{c|}{Method} &   L1 $\downarrow$  &   SSIM $\uparrow$  & PSNR $\uparrow$ &   L1 $\downarrow$  &   SSIM $\uparrow$  & PSNR $\uparrow$  &   L1 $\downarrow$  &   SSIM $\uparrow$  & PSNR $\uparrow$   &   L1 $\downarrow$  &   SSIM $\uparrow$  & PSNR $\uparrow$  \\
		\hline
		CPNet \cite{lee2019copy} & 0.011 & 0.95 & 32.25 & 0.0125 & 0.93 & 32.40 & 0.014 & 0.92 & 31.69 & 0.018 & 0.89 & 30.20 \\
		STTN \cite{zeng2020learning} & 0.011 & 0.94 & 32.63 & 0.013 & 0.93 & 32.08 & 0.017 & 0.90 & 30.99 & 0.024 & 0.83 & 28.40 \\
		DFG  \cite{xu2019deep}   & 0.008 & 0.94 & 33.12 & 0.008 & 0.95 & 33.27 & 0.008 & 0.96 & \textbf{33.37} & 0.009 & 0.95 & 32.37 \\
		FGVC \cite{gao2020flow}  & 0.006 & 0.95 & 33.20 & 0.006 & 0.95 & 33.20 & 0.006 & \textbf{0.97} & 33.20 & 0.006 & \textbf{0.97} & \textbf{33.20} \\
		Ours            & \textbf{0.005} &\textbf{0.97} & \textbf{34.66} & \textbf{0.002} & \textbf{0.97} & \textbf{34.66} & \textbf{0.003} & \textbf{0.97} & 33.14 & \textbf{0.003} & \textbf{0.97} &  33.14\\
	\end{tabular}
	\vspace{-8pt}
	\caption{Quantitative comparisons on the $\text{Syn-DS}^{+}$ dataset at various resolutions.}
	\label{table:comp}
	\vspace{-16pt}
\end{table*}

\vspace{-10pt}
\section{Experiments}
\vspace{-6pt}
In this section, we conduct quantitative and qualitative evaluations as well as ablation studies of our method on various datasets.
Our method produced superior results than the state-of-the-art methods for high resolution video inpainting in most cases. 

\vspace{2pt}
\noindent \textbf{Testing dataset.}
We test our method for inpainting various unseen videos after removing undesired objects or human specified by given masks.
First, we chose 50 sequences from the commonly used DAVIS dataset \cite{Perazzi2016} which provides high resolution videos with foreground object masks. 
As the original DAVIS object masks do not contain shadows, to make the inpainting result more realistic (without ghost shadows),
we added the shadows to the object masks using our in-house annotation tool.
We denote the shadow-annotated dataset as $\text{DS}^{+}$.
All $\text{DS}^{+}$ videos are resized to 1080p (1920x1080).
In addition, we collected 30 real world human-focused 1080p videos from the Internet and other datasets (denoted as HIN) and annotated some masks for undesired human in each video.
To enable the quantitative evaluation, we also generated a synthetic dataset ($\text{Syn-DS}^{+}$) by imposing the $\text{DS}^{+}$ masks onto another 50 real world 1080p videos which contain various natural scenes and city views.
 

\vspace{-20pt}
\subsection{Comparisons with Existing Methods}
\vspace{-4pt}

\rev{\textbf{Baselines.}}
We compare our STRA-Net\footnote{Our code and the improved masks which include the shadow annotation on top of the original DAVIS masks are available at
\url{https://github.com/Ascend-Research/STRA_Net}.
The code has been tested on GPU (CUDA) and Huawei Ascend processor (CANN).}
with the state-of-the art learning-based methods CPNet \cite{lee2019copy}, STTN \cite{zeng2020learning} and flow-based methods DFG \cite{xu2019deep}, FGVC \cite{gao2020flow}.
\rev{All experiments are done on a Tesla V100 GPU (32GB memory)} using the pre-trained models downloaded from their websites.
\rev{To verify the effectiveness of our different modules and the efficiency of our pipeline, we compare with the baselines on different video resolutions, ranging from low to high.
Since the 1080p videos cannot be supported by certain baselines such as STTN on a single 32GB GPU due to the memory constraint, we resized the 1080p videos in our testing dataset to various lower resolutions.
Proper padding is added to each resolution to fit the requirements of the baseline methods.
Finally, we obtain four video resolutions as shown in Table \ref{table:comp}.
}
As our model is trained with 512x512 videos, after padding, only temporal and spatial aggregation (while no residual aggregation) are performed for the low resolution 432x240 videos.
\rev{
Hence, the results on the 432x240 videos can be used to evaluate the performance of STA-Net alone instead of the full STRA-Net.
For all other resolutions, our STRA-Net will downsample the input video to 1/4 of the input resolution and run the STA-Net on the downsampled frames.
}

\noindent \textbf{Quantitative comparison.} 
\rev{As the $\text{Syn-DS}^{+}$ contains the ground truth frames after the synthetic $\text{DS}^{+}$ masks are removed,}
we conduct quantitative evaluation at each resolution of the resized $\text{Syn-DS}^{+}$ by computing the mean value of the standard L1, SSIM \cite{sampat2009complex} and PSNR \cite{korhonen2012peak} metrics between the completed and ground truth frames.
From Table \ref{table:comp}, our method produces superior results than the learning-based and flow-based baselines in most cases.

\rev{To further verify the efficiency of our pipeline}, we also compare the running time \rev{and memory consumption} for each method.
On average, to run on a 1728x960 frame, our STRA-Net takes 1.2 seconds which is similar to the CPNet (1s) and much faster than STTN (4s), DFG (48s) and FGVC (55s), showing the efficiency of our method.
\rev{For memory consumption, we compare with the baselines at two resolutions and our full pipeline (with the residual aggregation enabled) reports the lowest memory in both cases (Table \ref{table:memory}).
It can also be observed that the memory of other methods grows much faster than ours since we can always downsample the input video to a lower resolution (e.g., 1/4 of input) for running the STA-Net.
Note that the spatial and temporal residual aggregation can run on the CPU efficiently. 
Also, we didn't perform any memory optimization for all methods.
}

\begin{figure}[t]
	\centering
	\includegraphics[width=0.72\linewidth]{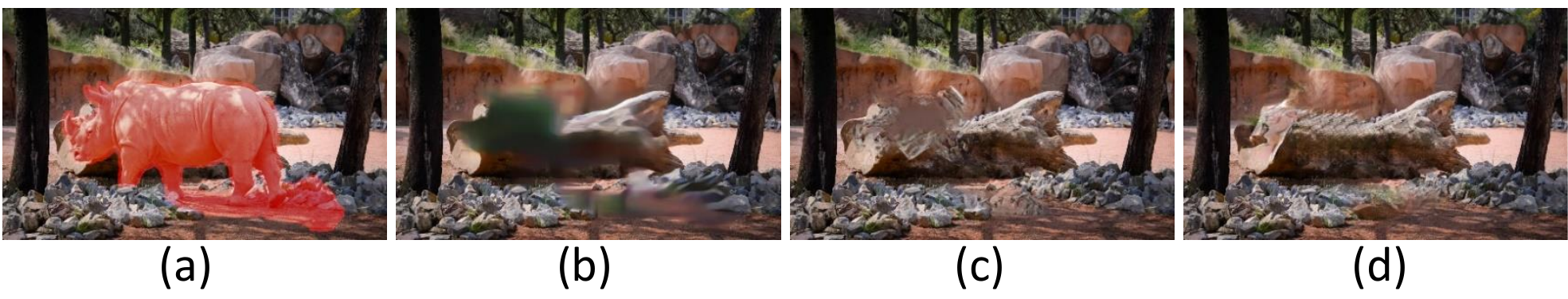}
	\vspace{-12pt}
	\caption{Video inpainting results on the $\text{DS}^{+}$ dataset (1728x960): (a) Input frame with mask (b) STTN (c) FGVC (d) Ours. }
	\label{fig:compare_davis}
	\vspace{-8pt}
\end{figure}

\begin{figure*}[t]
	\centering
	\includegraphics[width=0.98\linewidth]{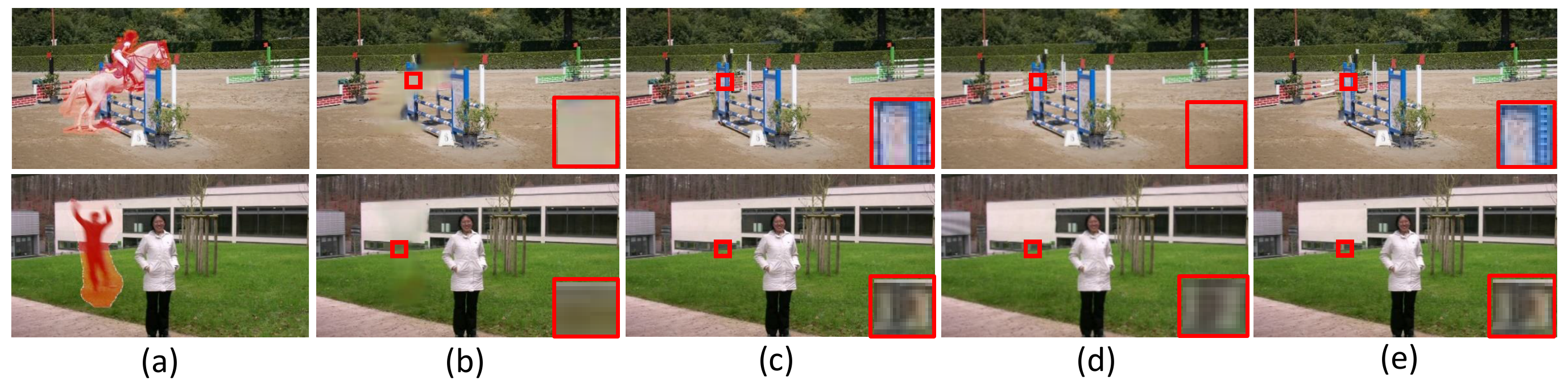}
	\vspace{-12pt}
	\caption{Ablation study for temporal modules on $\text{DS}^{+}$ (first row) and \revsupp{HIN (second row)}: (a) Input frame with mask (1080p) (b) Vanila temporal aggregation (270p) (c) Multi-scale temporal aggregation (270p) (d) After bilinear upsampling (1080p) (e) With temporal residuals added to (d). }
	\label{fig:temp_ablation}
	\vspace{-10pt}
\end{figure*}

\begin{table*}
    \vspace{-2pt}
	\centering
		\footnotesize
	\addtolength{\tabcolsep}{-4pt}
	\begin{tabular}{l|cccc| cccc } &
 \multicolumn{4}{c|}{864 $\times$ 480} &   \multicolumn{4}{c}{1728 $\times$ 960} \\
		
		\multicolumn{1}{c|}{Method} &   CPNet \cite{lee2019copy}  &   STTN \cite{zeng2020learning} & FGVC \cite{gao2020flow} &  Ours  &    CPNet \cite{lee2019copy}  &   STTN \cite{zeng2020learning} & FGVC \cite{gao2020flow} &  Ours  \\
		\hline
		Memory (GB) & 3.3 & 5.7 & 2.6 & \textbf{2.0} 
		& 10.1 & 20.6 & 18.7 & \textbf{3.7}\\
	\end{tabular}
	\vspace{2pt}
	\caption{\rev{Comparison on the memory consumption of different methods.}}
	\label{table:memory}
	\vspace{-8pt}
\end{table*}

\noindent \textbf{Qualitative comparison.} We compare different methods on both real world videos in $\text{DS}^{+}$, HIN and the synthetic ones in $\text{Syn-DS}^{+}$. 
In Figure \ref{fig:compare_davis}, our method produces more meaningful inpainting than other methods on $\text{DS}^{+}$.
Due to the page limit, more results on HIN and $\text{Syn-DS}^{+}$ can be found in the supplementary material.
Overall, our method can generate more temporally and spatially consistent results due to the proposed joint image alignment as well as the attention-guided multi-scale temporal and spatial aggregation.

\noindent \textbf{User study.} \rui{We conduct a user study by asking 15 volunteers from the computer science domain to rank the inpainting results from our method along with CPNet, STTN and FGVC.
We randomly chose results (generated with 1728x960 input) of 20 sequences from $\text{DS}^{+}$ and 10 sequences from HIN to make a study of 30 tasks.
From the study results, our method is ranked as the best in 39.8\% cases, and the second best in 53.6\% cases, comparing to the 54.9\% and 30.0\% of FGVC to be ranked as the best and the second.
Note that for some results, the top two results are actually very similar. 
For these cases, we ask the user to randomly choose one as the best. 
Based on the user study, although our method got slightly inferior results than FGVC with some randomness, it performs far better than CPNet and STTN.
Also, FGVC can take much longer time to compute the optical flow for high resolution videos, while our method runs much faster and requires much less memory since our STA-Net only runs on low resolution frames.
More details of the user study setup and results are provided in the supplementary material. 
}

 \begin{figure*} [t]
 	\centering
 	\includegraphics[width=0.62\linewidth]{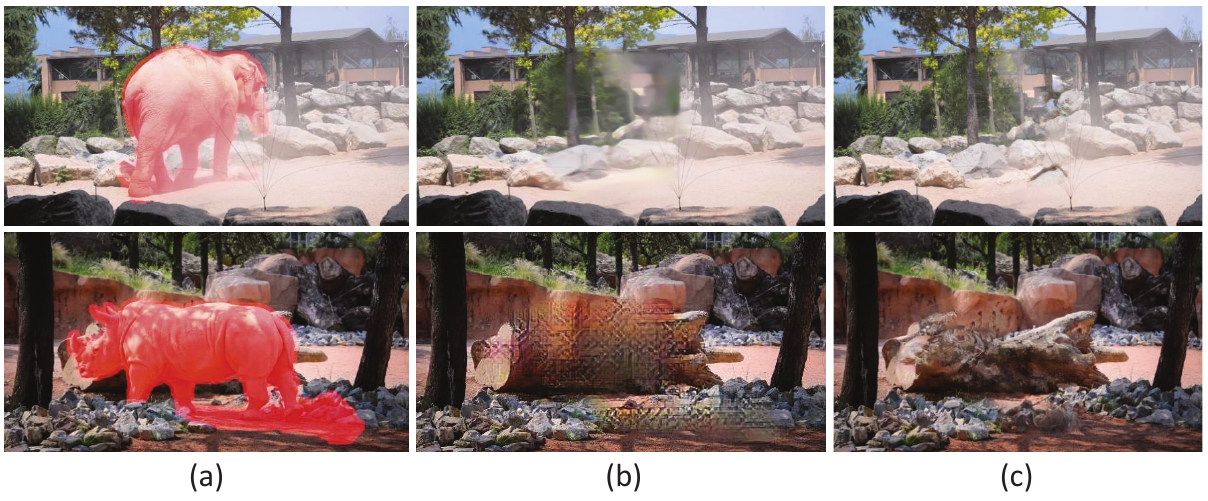}
 	\vspace{-10pt}
 	\caption{\revsupp{Ablation study for the spatial aggregation module on $\text{DS}^{+}$. (a) Input frame with mask (270p) (b) Without spatial aggregation (270p) (c) With spatial aggregation (270p).}}
 	\label{fig:spatial_aggregation}
 	\vspace{-14pt}
 \end{figure*}


\begin{figure}[t]
	\centering
	\includegraphics[width=0.75\linewidth]{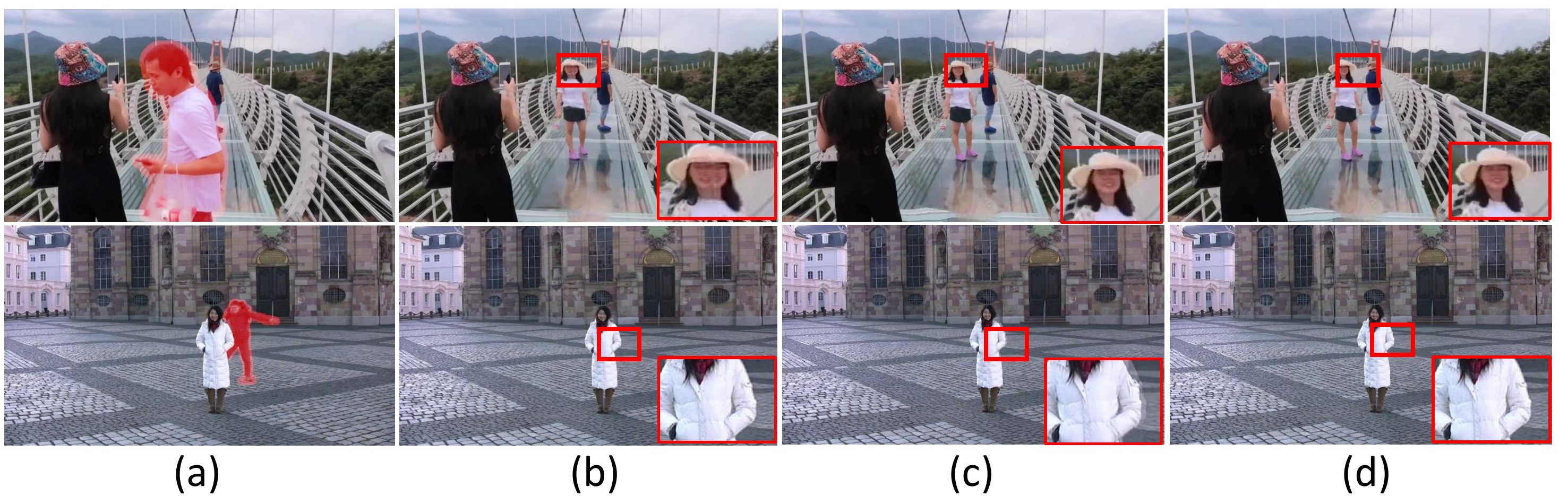}
	\vspace{-12pt}
	\caption{Ablation study for different image alignment methods on HIN: (a) Input frame with mask (1080p); (b) Affine alignment only; (c)  Flow alignment only; (d) Both. 
		Flow alignment works better in the first row and affine alignment works better in the second. The joint alignment works well in both cases.}
	\label{fig:alignment_ablation}
	\vspace{-10pt}
\end{figure}

\vspace{-6pt}
\subsection{Ablation Studies}
\vspace{-4pt}
\rev{
We perform extensive ablation studies to identify the effectiveness of the proposed modules.
As our method can naturally support high resolution videos, all ablation studies are conducted on 1080p videos unless the resolution is specified.
The corresponding downsampled low resolution videos are 270p (480x270).
}

\label{sec:ablation}
\noindent \textbf{Analysis of different spatial and temporal modules.}
We conduct quantitative ablation studies for different spatial and temporal modules on the $\text{Syn-DS}^{+}$ dataset.
The results in Table \ref{table:ab} show the contributions of individual components.  
 We also visualize the effects of different temporal modules in Figure \ref{fig:temp_ablation} on the real world $\text{DS}^{+}$ videos.
 \revsupp{
 The proposed multi-scale temporal aggregation and the temporal residual aggregation can
generally improve the details in the inpainting result comparing to the vanila baseline.
 In Figure \ref{fig:spatial_aggregation}, we can see the contribution of our multi-scale spatial aggregation for improving the inpainting quality on the central regions where no information has been transferred from other temporal reference frames, e.g., the region covered by the leftover mask. 
 }
 \rui{Overall, the results show our proposed modules are essential to improve the performance from the baselines.}

\begin{table*}
    \vspace{-1pt}
	\centering
	\footnotesize
	\addtolength{\tabcolsep}{-5pt}
	\begin{tabular}{l|ccc || l| ccc}
		&  \multicolumn{3}{c||}{432 $\times$ 240} &  &\multicolumn{3}{c}{1728 $\times$ 960}  \\
		
		\multicolumn{1}{c|}{Module} &   L1 $\downarrow$  &   SSIM $\uparrow$  & PSNR $\uparrow$ &  \multicolumn{1}{c|}{Module}& L1 $\downarrow$  &   SSIM $\uparrow$  & PSNR $\uparrow$   \\
		\hline
		Vanila temporal aggregation   & 0.024 & 0.85 & 28.36 & Bilinear up-sampling   & 0.033 & 0.80 & 26.39\\
Multi-scale temporal aggregation   & 0.005 & \textbf{0.97} & 34.66 & Temporal residual aggregation   & 0.005 & 0.96 & 33.11 \\
+ Multi-scale spatial aggregation   & \textbf{0.003} & \textbf{0.97} & \textbf{34.89} & + Spatial residual aggregation   & \textbf{0.003} & \textbf{0.97} & \textbf{33.14}
	\end{tabular}
\vspace{2pt}
	\caption{Quantitative ablation study of different modules on $\text{Syn-DS}^{+}$. Left:  multi-scale spatial and temporal aggregation modules on low resolution videos (no spatial or temporal residual aggregation is applied). Right: spatial and temporal residual aggregation modules on high resolution videos (both multi-scale temporal and spatial aggregation are enabled). }
	\label{table:ab}
	\vspace{-8pt}
\end{table*}

%
%
%
%

 \begin{table}[!t]
 	\centering
 	\footnotesize
 	\addtolength{\tabcolsep}{-3pt}
 	\begin{tabular}{l | ccc }
 		\multicolumn{1}{c|}{Alignment Method}  &   L1 $\downarrow$  &   SSIM $\uparrow$  & PSNR $\uparrow$ \\
 		\hline
 		Affine transformation & 0.005 & \textbf{0.98} & 32.05\\
 		Optical flow  & 0.017 & 0.927 & 29.21 \\
 		Both  & \textbf{0.003} & 0.97 & \textbf{33.14} \\		
 	\end{tabular}
 	\caption{\revsupp{Quantitative ablation study for different image alignment methods on $\text{Syn-DS}^{+}$. Results are generated at 1080p resolution.}}
 	\label{table:alignment_abl}
 	\vspace{-12pt}
 \end{table}

\vspace{2pt}
\noindent \textbf{Analysis of alignment method.}
We conduct an ablation study to evaluate the performance of different image alignment methods. 
As shown in Figure \ref{fig:alignment_ablation},  \rev{affine alignment which can utilize global information and temporal features from far away frames produces better results if the mask region is small or if there is global motion like camera shake/pan involved. 
On the other hand, flow alignment works better when the mask region is large and the alignment can mainly be inferred based the local movements.}
Hence, both methods can excel in certain situations.
 Our final joint alignment module combines the strengths of both methods and achieves consistent results in different cases.
\revsupp{
In Table \ref{table:alignment_abl}, quantitative results verify that our joint alignment method can generally improve the performance of each individual method.
}

\vspace{-10pt}
\section{Conclusion}
\vspace{-2pt}
We present STRA-Net, a novel and modularized learning-based framework for using spatial-temporal residual aggregation to achieve high resolution video inpainting.
We believe the STRA-Net can be easily adapted to other video inpainting methods to improve their performance on high resolution videos.
In the future, we would like to explore its application in other video processing tasks such as efficient HD video synthesis and animation \cite{yi2020animating}.

\section{Acknowledgement}
We thank Qian (Derek) Wang for providing assistance on open sourcing the code . 
We also thank the anonymous reviewers who have provided valuable feedback to improve the paper.

\bibliography{references}

\appendix

\section{Appendix}

\subsection{Loss Functions}
\rev{
The definition of loss functions are similar as in CPNet \cite{lee2019copy} and HiFill \cite{yi2020contextual}.
The main difference is how the output of the main steps is calculated, e.g., we use the joint alignment while CPNet only uses the affine alignment. Also, our multi-scale spatial aggregation only applies on the leftover mask region instead of the full mask as in HiFill.
Also, note that the loss functions are used to train the STA-Net which works on the downsampled low resolution frames.
The specific definitions are as follows:
\begin{equation}
L_{align} = \frac{1}{N} \cdot {\sum_{t}^{N} \sum_{r}  \mid \mid (1-M^{t}) \odot (1-M^{r \rightarrow t}) \odot (X^{t} -X^{r \rightarrow t}) \mid \mid _1}
\label{eq:align}
\end{equation}

\vspace{-12pt}

\begin{equation}
L_{hole(visible)} = \frac{1}{N} \cdot {\sum_{t}^{N}  \mid \mid M^{t} \odot C_{visible} \odot (Y^{t} -Y_{GT}^{t} ) \mid \mid _1} 
\label{eq:vis}
\end{equation}

\vspace{-10pt}

\begin{equation}
L_{hole(leftover)} = \frac{1}{N} \cdot {\sum_{t}^{N} \mid \mid M_{leftover}^t \odot (Y^{t} -Y_{GT}^{t} ) \mid \mid _1
\label{eq:leftover}} 
\end{equation}

\vspace{-8pt}

\begin{equation}
L_{non-hole} = \frac{1}{N} \cdot {\sum_{t}^{N}  \mid \mid (1 - M^t )\odot (Y^{t} - Y_{GT}^{t} ) \mid \mid _1
\label{eq:non-hole}} 
\end{equation}

\vspace{-8pt}

\begin{equation}
L_{perceptual} = \frac{1}{N} \cdot {\sum_{t}^{N} \frac{1}{P} \cdot \sum_{p}^{P}  \mid \mid \phi_p(Y_{comb}^{t}) - \phi_p(Y_{GT}^{t})  \mid \mid _1} 
\label{eq:percep}
\end{equation}

\vspace{-8pt}

\begin{equation}
L_{style} = \frac{1}{N} \cdot {\sum_{t}^{N} \frac{1}{P} \cdot \sum_{p}^{P}  \mid \mid G^{\phi}_p(Y_{comb}^{t}) - G^{\phi}_p(Y_{GT}^{t})  \mid \mid _1} 
\label{eq:style}
\end{equation}

\vspace{-10pt}

\begin{equation}
L_{rec} = \frac{1}{N} \cdot {\sum_{t}^{N}  \mid \mid M_{leftover}^t \odot (\tilde{X}^t - X^t)  \mid \mid _1 
+  \mid \mid (1- M_{leftover}^t) \odot ( \tilde{X}^t - X^t) \mid \mid _1} 
\label{eq:rec}
\end{equation}

\vspace{-10pt}

\begin{equation}
L_{adv} = \frac{1}{N} \cdot {\sum_{t}^{N} -\mathbb{E}_{{\tilde{X}^t} \in \mathbb{P}_g}  [ D(\tilde{X}^t) ]}
\label{eq:adv}
\end{equation}



In above equations, $X^t$ is the downsampled input frame at time $t$ and $M^t$ is its corresponding input mask.
$Y^t$ is the final inpainting result of $X^t$ and $Y_{GT}^t$ is the ground truth for $Y^t$.
$N$ is the number of frames in each sample of the training videos.
For our current training data, $N=5$.
$\odot$ is the element-wise multiplication.
In Equation \ref{eq:align}, $X^{r \rightarrow t}$ and $M^{r \rightarrow t}$ is the aligned reference frame and its aligned mask computed by aligning the reference frame $X^r$ to $X^t$ using the joint alignment module.
In Equation \ref{eq:vis} and \ref{eq:leftover}, $C_{visible}$ is the aggregated temporal attention scores (see the main paper and \cite{lee2019copy} for the definition);
$M_{leftover}^t$ is the leftover mask for frame $t$ which is defined as $M_{leftover}^t = M^t \odot (1-C_{visible})$.
In Equation \ref{eq:percep} and \ref{eq:style}, $Y_{comb}^t$ is the combination of the inpainting output $Y^t$ in the hole region and the input $X^t$ outside the hole;
$p$ is the index of the pooling layer in VGG-16 \cite{Simonyan15_vgg} and $\phi_p(\cdot)$ is the output of the corresponding layer;
$G_p^{\phi}(\cdot)$ is the gram matrix
multiplication \cite{johnson2016perceptual}.
In Equation \ref{eq:rec} and \ref{eq:adv}, $\tilde{X}^t$ is the generator output which is defined as $\tilde{X}^t = G(X^t, M_{leftover}^{t})$, where $G(\cdot, \cdot)$ is the generator;
$D(\cdot)$ is the discriminator and $\mathbb{P}_g$ is the distribution of the input frames. Note that the adversarial training loss is designed in the same way as in \cite{yi2020contextual}. More details about defining the WGAN-GP loss for the multi-scale spatial aggregation can also be found in \cite{yi2020contextual}}

\subsection{More Training Details}
In this section, we provide more details about training the STA-Net. The network is trained
on the proposed synthetic training dataset which contains numerous short video clips and the
ground truth object masks. We train the model on 2 Tesla V100 GPUs for 10 epochs with
the batch size set to 40. 
In our experiments, we increase the weight of $L_{align}$ so that the alignment module gets trained first and the later training converges faster.
The joint alignment and temporal aggregation parts of the network
are warmed-up for 2 epochs before adding the spatial aggregation module to the network for
training stability reasons. The model is implemented in PyTorch.

\begin{figure*}[t]
\centering
	\includegraphics[width=0.95\linewidth]{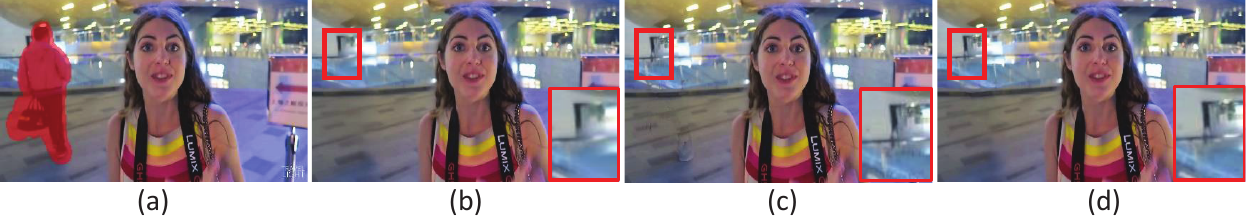} 
	\vspace{-10pt}
	\caption{Video Inpainting result on human-focused real world HIN dataset: (a) Input frame
with mask (1728x960) (b) STTN (c) FGVC (d) Ours.}
\label{fig:compare_vin}
	\vspace{-10pt}
\end{figure*}

   \begin{figure} [t]
	\centering
	\includegraphics[width=0.95\linewidth]{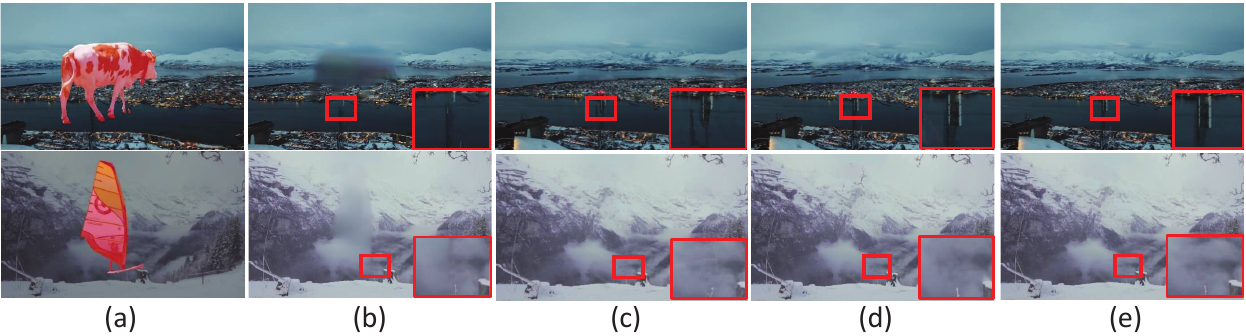}
	\vspace{-10pt}
	\caption{Video inpainting results on $\text{Syn-DS}^{+}$ dataset: (a) Input frame with mask (1728x960)
(b) STTN (c) FGVC (d) Ours (e) Ground Truth.}
	\label{fig:compare_syn}
	\vspace{-10pt}
\end{figure}

\subsection{More Experiment Results}
\paragraph{Qualitative comparison.}
In Figure \ref{fig:compare_vin} and \ref{fig:compare_syn}, we show more results for qualitative comparisons of our methods with the state-of-the-art learning and flow-based methods on the
synthetic $\text{Syn-DS}^{+}$ and real world HIN dataset. We also attach a demo video that compares
results from different methods for video inpainting on 1080p videos from the $\text{DS}^{+}$ \cite{Perazzi2016} and
the HIN dataset.
Note that some baselines cannot directly run on 1080p videos due to memory constraint.
Therefore, we run these baselines on 1792x960 resolution and then resize the output to 1080p.
The improvements of our method upon others can be observed more clearly
in the video. The results show that the learning based method STTN \cite{zeng2020learning} tends to generate
blurry inpaintings and the flow-based method FGVC \cite{gao2020flow} may produce artifacts near the mask
boundaries. Our results are more temporally coherent and visually appealing.

\begin{figure} [!t]
 	\centering
 	\includegraphics[width=0.85\linewidth]{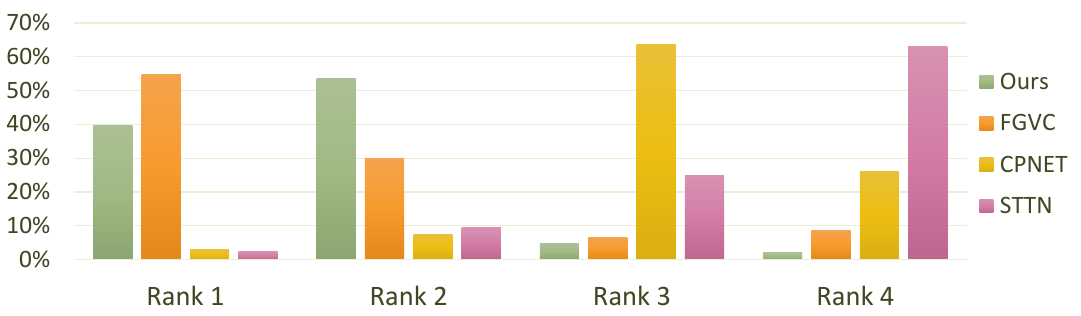}
 	\vspace{-10pt}
 	\caption{User study on 30 tasks composed of videos from $\text{DS}^{+}$ and HIN. “Rank x” means
the percentage of results from each model being chosen as the x-th best.}
 	\label{fig:user_study}
 	\vspace{-5pt}
 \end{figure}

\paragraph{User study.}
In the user study, there are 30 tasks composed of video inpainting results generated at 1728x960 resolution of
the $\text{DS}^{+}$ and HIN dataset . In each task, the input video with the masks is first shown to the user, then results of the four methods are randomly arranged. The user is asked to rank
the four results based on two criteria: 1) whether the inpainted videos look like real; 2)
whether the inpainting results are spatially and temporally consistent. Figure \ref{fig:user_study} summarizes
the user study results. It can be found that our results are comparable to the state-of-theart flow-based method FGVC while outperform the learning-based methods CPNet \cite{lee2019copy} and
STTN comfortably for the high resolution videos. Another important observation is our method has
the lowest percentage of being ranked as the last which shows the robustness of our method.

\end{document}